\def\year{2019}\relax
\documentclass[letterpaper]{article} 
\usepackage{aaai19}  
\usepackage{times}  
\usepackage{helvet}  
\usepackage{courier}  
\usepackage{url}  
\usepackage{graphicx}  
\usepackage[utf8]{inputenc} 
\usepackage[T1]{fontenc}    
\usepackage{url}            
\usepackage{booktabs}       
\usepackage{amsfonts}       
\usepackage{nicefrac}       
\usepackage{microtype}      
\usepackage{mathtools}
\usepackage{amsmath, amssymb}
\usepackage[capitalize]{cleveref}
\usepackage{subcaption}
\usepackage{algorithm}
\usepackage{algorithmic}
\usepackage{color}

\DeclareMathOperator*{\E}{\mathbb{E}}

\DeclareMathOperator*{\argmax}{arg\:max}

\newcommand{\data}{x}
\newcommand{\Data}{X}
\newcommand{\truelabel}{y}
\newcommand{\truelabels}{\boldsymbol{y}}
\newcommand{\mylabel}{\hat{y}}
\newcommand{\labels}{\hat{\truelabels}}
\newcommand{\numdata}{n}
\newcommand{\numweak}{m}
\newcommand{\weakvec}{\boldsymbol{q}}
\newcommand{\bounds}{\boldsymbol{b}}
\newcommand{\bound}{b}
\newcommand{\learnedprob}{p}
\newcommand{\learnedprobs}{\boldsymbol{\learnedprob}}
\newcommand{\function}{f}
\newcommand{\params}{\theta}
\newcommand{\primal}{g}
\newcommand{\lagrange}{\gamma}
\newcommand{\lagrangevec}{\boldsymbol{\lagrange}}
\newcommand{\lagrangian}{L}
\newcommand{\stepsize}{\rho}
\newcommand{\rate}{\alpha}
\newcommand{\rates}{\boldsymbol{\alpha}}
\newcommand{\placeholder}{\boldsymbol{z}}

\newcommand{\reals}{\mathbb{R}}

\begin{document}
%
\title{Adversarial Label Learning}
\author{
Chidubem Arachie \\
  Department of Computer Science\\
  Virginia Tech \\
  achid17@vt.edu \\
\And
Bert Huang\\
  Department of Computer Science\\
  Virginia Tech \\
  bhuang@vt.edu
  }
\maketitle
\begin{abstract}
We consider the task of training classifiers without labels. We propose a weakly supervised method---adversarial label learning---that trains classifiers to perform well against an adversary that chooses labels for training data. The weak supervision constrains what labels the adversary can choose. The method therefore minimizes an upper bound of the classifier's error rate using projected primal-dual subgradient descent. Minimizing this bound protects against bias and dependencies in the weak supervision. Experiments on real datasets show that our method can train without labels and outperforms other approaches for weakly supervised learning. 
\end{abstract}

\section{Introduction}

This paper introduces \emph{adversarial label learning} (ALL), a method for training classifiers without labels by making use of weak supervision. ALL works by training classifiers to perform well on adversarially labeled instances that are consistent with the weak supervision. Many machine learning models require large amounts of labeled training data, which is usually hand labeled or observed and recorded. In real applications, large amounts of training data are often not easily accessible or are expensive to acquire, making labeled training data a critical bottleneck for machine learning. 

An alternative for training machine learning models without labeled training data is \emph{weak supervision}. Weak supervision uses domain knowledge about the specific problem, side information, or heuristics to approximate the true labels. A key challenge for weak supervision is the fact that there may be bias in the errors made by the weak supervision signals. Using multiple sources of weak supervision can somewhat alleviate this concern, but dependencies among these weak supervision functions can be misconstrued as independent confirmation of erroneous labels. 
For example, in a classification task to identify diabetic patients, physicians know that obesity can indicate diabetes, and they also know the rate at which this indicator is wrong. However, since the indicator is biased, models trained with this information will learn to detect obesity, not the original goal of diabetes. To correct this problem, one may also consider high blood pressure as a second weak indicator. Unfortunately, these indicators are correlated and may make dependent errors. 

ALL trains using weak supervision and aims to mitigate these problems by adversarially labeling the data. The adversarial labeling can construct scenarios where dependencies in the weak supervision are as confounding as possible while preserving the partial correctness of the weak supervision. The learner then trains a model that can perform well against this adversarial labeling. 
ALL solves these two competing optimizations using primal-dual subgradient descent. The inner optimization finds a worst-case distribution of the labels for the current weight parameter of the model, while the outer optimization finds the best weights for the model for the current label distribution. The inner optimization's maximized error rate can also be viewed as an upper bound on the true error rate, which the outer optimization aims to minimize. By training to perform well on the worst-case labeling, ALL is robust against dependent and biased errors in weak supervision signals.

The inputs to ALL are a set of unlabeled data examples, a set of weak supervision signals that approximately label the data, and a corresponding set of estimated error bounds on these weak supervision signals. Domain experts can design the weak supervision signals---e.g., by defining approximate labeling rules---and they can use their knowledge to set bounds on the errors of these signals. When designing weak supervision signals, experts often have mental estimates of how noisy the signals are, so this error estimate is an inexpensive yet valuable input for the learning algorithm. 

We consider a binary classification setting where a parameterized model is trained to classify the data. We make use of multiple weak signals that represent different approximations of the true model. These weak signals can be interpreted as having different views of the data. The estimated error rates of these weak signals are passed as constraints to our optimization. Importantly, we show that ALL works in cases where these weak signals make dependent errors. Our experiments also show that ALL trains classifiers that are better than the weak supervision signals, even when the error estimates are incorrect. The performance of ALL in this setting is significant because domain experts will often imperfectly estimate the noisiness of the weak supervision signals.  


\section{Related Work}
\label{sec:related}

Weak supervision has become an important topic in the context of data-hungry deep learning models. A new line of research on data programming has produced a paradigm for weak supervision where data scientists write labeling functions that create noisy labels  \cite{ratner2017snorkel,ratner2016data}. The approach then discovers relationships among the noisy labeling functions and is able to combine them and train data-hungry models. Other related approaches provide weak supervision in the form of constraints on the output space \cite{stewart2017label}, such as those that encode physical laws. 
Another related effort is on meta-learning for neural networks via weak supervision \cite{dehghani2017learning}, using semi-supervised data to train an algorithm to learn from weak supervision. 

Our work is related to existing methods that use variants of a generalized expectation (GE) criteria \cite{druck2008learning,mann2010generalized,mann2008generalized} for semi- and weakly supervised learning. A GE criterion \cite{mccallum2007generalized} is a term in a parameter estimation objective function that prefers models to match conditional probabilities provided as weak supervision. 
These conditional probabilities may take the form of the probability of labels given a feature \cite{druck2008learning}, also allowing the weak supervision to include information about the uncertainty of a weak signal. 
Posterior regularization (PR) \cite{ganchev2010posterior} is a similar approach that trains models to adhere to constraints on their output posterior distributions. These constraints can also take the form of weak supervision signals that specify the class of allowable posterior distributions for the learned model.
While GE and PR allow incorporation of weak supervision and quantification of weak signal errors, they do not explicitly consider that these weak signals may make errors that conspire to confound the learner. 
Our development of ALL aims to address this shortcoming.

Our work is also related to methods developed to estimate the error of classifiers without labeled data \cite{jaffe2016unsupervised,platanios2014estimating,steinhardt2016unsupervised} that rely on statistical relationships between the error rates of different classifiers. Many of these approaches extend classical statistics methods \cite{dawid1979maximum} by allowing the errors of the different classifiers to be dependent variables. A key goal of these approaches is to infer the error rate of these classifiers given only unlabeled data. In contrast, our setting assumes that we have reasonably good estimates of the error rates for the weak supervision provided by experts. 

A different form of adversarial learning has recently become popular for deep learning \cite{goodfellow2014generative}. Generative adversarial networks (GANs) pit a data generator and a discriminator against each other to train generative models that imitate realistic data distributions. Though our goal is not to train generative models, the stochastic optimization techniques developed for GANs may help our future work. \emph{Virtual adversarial training} \cite{miyato2018virtual} uses \emph{input} perturbation to regularize a semi-supervised learning method. The method adds a regularization term to the objective function to make the learned model robust to input perturbations. Other approaches on adversarial input perturbation include methods for adversarial training of structured predictors \cite{torkamani2013convex,torkamani2014robustness}, which lead to the added benefit of generalization guarantees. Our approach focuses on adversarial output manipulation, and opportunities to combine the benefits of both are promising directions of future work.

Other research \cite{lowd2005adversarial,madry2017towards} has considered variants of adversarial learning, training a classifier to learn sufficient information about another classifier to construct adversarial attacks. These efforts primarily focus on training models to be robust against malicious attacks, which is of interest in cybersecurity. 

\section{Adversarial Label Learning}
\label{sec:method}

The principle behind adversarial label learning (ALL) is that we train a model to perform well under the worst possible conditions. The conditions being considered are the possible labels of the training data. We consider the setting in which the learner has access to a training set of examples,
and weak supervision is given in the form of some approximate indicators of the target classification along with expert estimates of the error rates of these indicators. Formally, let the data be $\Data = [\data_1, \ldots, \data_\numdata ]$. (We consider these examples to be ordered for notational convenience, but the order does not matter.) These examples belong to classes $[\truelabel_1, \ldots, \truelabel_\numdata ] \in \{0, 1\}^\numdata $. The training labels $\truelabels$ are unavailable to the learner. Instead, the learner has access to $\numweak$ weak supervision signals $\{\weakvec_1, \ldots, \weakvec_\numweak\}$, where each weak signal is a soft labeling of the data, i.e., $\weakvec_i \in [0, 1]^\numdata$. These soft labelings are estimated probabilities that the example is in the positive class. In conjunction with the weak signals, the learner also receives estimated expected error rate bounds of the weak signals $\bounds = [\bound_1, \ldots, \bound_\numweak]$. These values bound the expected error of the weak signals, i.e., 
\begin{equation}
\textstyle{
\bound_i \ge \E_{\labels \thicksim \weakvec_i} \left[ \frac{1}{\numdata} \sum_{j=1}^\numdata \left[ \mylabel_j \neq \truelabel_j \right] \right]~,}
\end{equation}
which can be equivalently expressed as
\begin{equation}
\textstyle{\bound_i \ge \frac{1}{\numdata} \left( \weakvec_i ^\top (1 - \truelabels) + (1 - \weakvec_i)^\top \truelabels \right)~.}
\label{eq:bound}
\end{equation}

While the learned classifier does not have access to the true labels $\truelabels$, it will use the assumption that this bound holds to define the space of possible labelings. Let the current estimates of learned label probabilities be $\learnedprobs \in [0, 1]^\numdata$. We relax the space of discrete labelings to the space of independent probabilistic labels, such that the value $\mylabel_j \in [0, 1]$ represents the probability that the true label $\truelabel_j$ of example $\data_j$ is positive. 
The adversarial labeling then is the vector of class probabilities $\labels$ that maximizes the expected error rate of the learned probabilities subject to the constraints given by the weak supervision signals and bounds, which can be found by solving the following linear program:
\begin{equation}
\begin{aligned}
\argmax_{\labels \in [0, 1]^\numdata} ~~ & ~~ \tfrac{1}{\numdata} \left( \learnedprobs ^\top (1 - \labels) + (1 - \learnedprobs)^\top \labels \right)\\
\textrm{s.t.} ~~ & ~~ \bound_i \ge \tfrac{1}{\numdata} \left( \weakvec_i ^\top (1 - \labels) + (1 - \weakvec_i)^\top \labels \right), \\
~ \forall i \in \{1, \ldots, \numweak\} ~ ,
\end{aligned}
\end{equation}
which we present in this unsimplified form to convey the intuition behind its objective and constraints; some algebra simplifies this optimization into a more standard form.

The adversarial labeling described so far is a key component of the learning algorithm. ALL trains a parameterized prediction function $\function_\params$ that reads the data as input and outputs estimated class probabilities, i.e., $\left[\function_\params(\data_j)\right]_{j=1}^\numdata =  \learnedprobs$. We will write $\learnedprobs(\params)$ to mean $\left[\function_\params(\data_j)\right]_{j=1}^\numdata$ when it is important to note that these are generated from the parameterized function $\function$. For now, we assume a general form for this parameterized function. For our optimization method described later in \cref{sec:optimization}, we assume that the function $\function$ is sub-differentiable with respect to its parameters $\params$. The goal of learning is then to minimize the expected error relative to the adversarial labeling. This principle leads to the following saddle-point optimization:
\begin{equation}
\begin{aligned}
\min_{\params} ~~ \max_{\labels \in [0, 1]^\numdata} ~~ & ~~ \tfrac{1}{\numdata} \left( \learnedprobs(\params) ^\top (1 - \labels) + (1 - \learnedprobs(\params))^\top \labels \right)\\
\textrm{s.t.} ~~ & ~~ \bound_i \ge \tfrac{1}{\numdata} \left( \weakvec_i ^\top (1 - \labels) + (1 - \weakvec_i)^\top \labels \right), \\
~ \forall i \in \{1, \ldots, \numweak\} ~ .
\end{aligned}
\label{eq:mainobjective}
\end{equation}

We can view the outer optimization as optimizing a primal objective that is the maximum of the constrained inner optimization. Define this primal function as $\primal(\params)$, such that \cref{eq:mainobjective} can be equivalently written as $\min_\theta \primal(\params)$. 
If the weak supervision error bounds are true, \emph{this primal objective value is an upper bound on the true error rate}. This fact can be proven by considering that the true labels $\truelabels$ satisfy the constraints, and the inner optimization seeks a labeling $\labels$ that maximizes the classifier's expected error rate. In the next section, we visualize this primal function and the behavior of adversarial labeling before describing how we efficiently solve this optimization in \cref{sec:optimization}.

\subsection{Visualizing Adversarial Label Learning}

In this section, we investigate a simple case that illustrates the behavior of the primal objective function $\primal$ on a two-example dataset ($\numdata = 2$). For a small dataset, we can visualize in two dimensions a variety of concepts.

\begin{figure}[tbh]
    \centering
    \begin{subfigure}[b]{0.35\textwidth}
        \includegraphics[width=\textwidth]{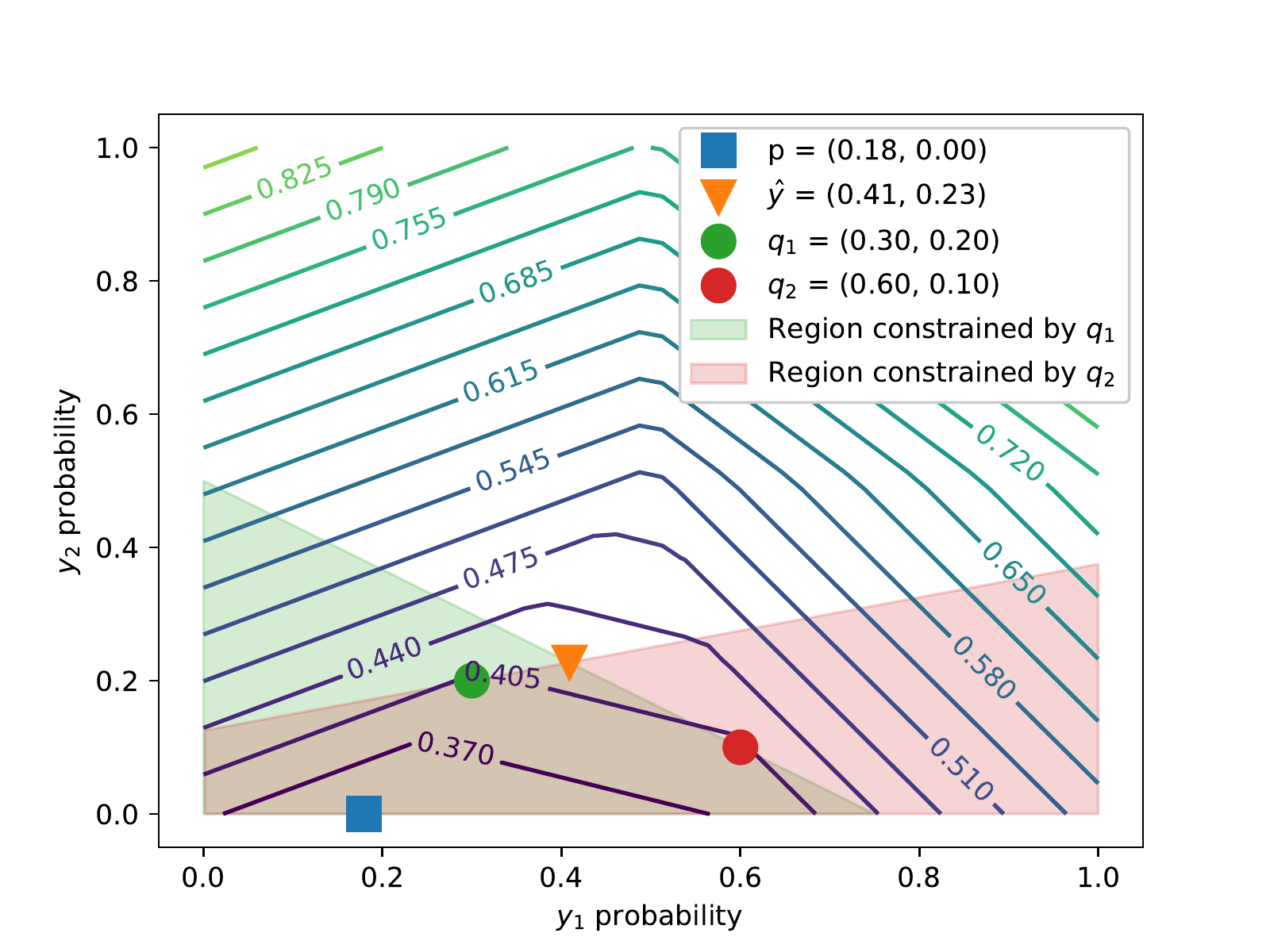}
        \caption{Two weak signals}
        \label{fig:optimal}
    \end{subfigure}
    \begin{subfigure}[b]{0.35\textwidth}
        \includegraphics[width=\textwidth]{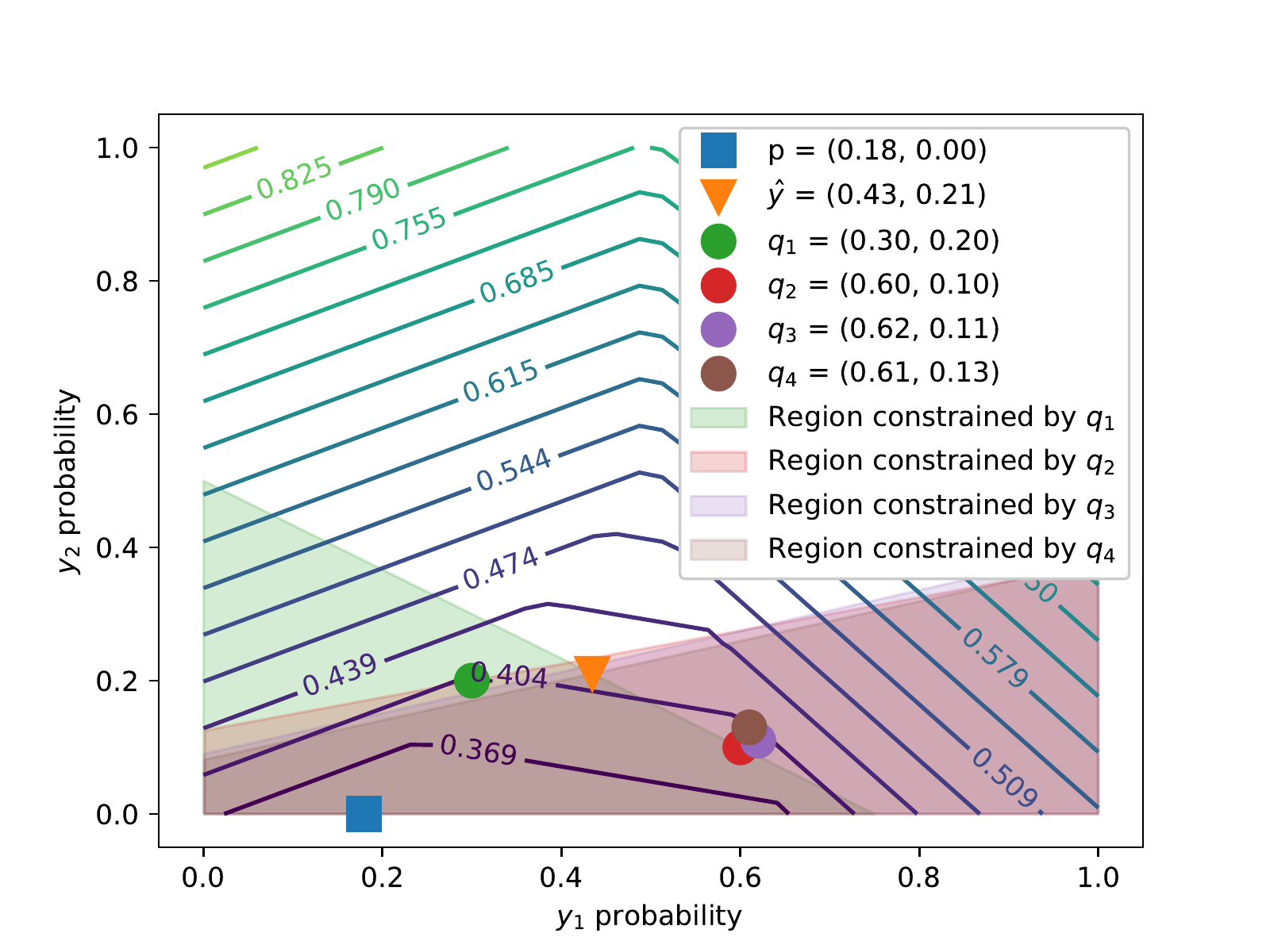}
        \caption{Redundant weak signals}
        \label{fig:extraweak}
    \end{subfigure}
    \caption{Illustrations of the primal objective function from \cref{eq:mainobjective}, the constraints set by the weak supervision, and the optimal learned probabilities and adversarial labels for a two-example problem.}\label{fig:2dviz}
\end{figure}

In \cref{fig:optimal}, we illustrate the constraints set by the two weak supervision signals. The first signal $\weakvec_1$ estimates that $\mylabel_1$ is positive with probability 0.3 and that $\mylabel_2$ is positive with probability 0.2. The second signal $\weakvec_2$ estimates that $\mylabel_1$ is positive with probability 0.6 and that $\mylabel_2$ is positive with probability 0.1. The bounds for each weak signal error are set to $\bound_1 = \bound_2 = 0.4$. Note that both weak signals agree that $\mylabel_2$ is most likely negative, but they disagree on whether $\mylabel_1$ is more likely to be positive or negative. 

\noindent\textbf{Constraints on $\labels$} ~~ The shaded regions represent the feasible regions determined by the linear constraint corresponding to each weak signal. The intersection of these feasible regions is the search space for label vectors. Note how the pink region determined by $\weakvec_2$ allows $\mylabel_1$ to be either extreme of 0 or 1. With more examples ($\numdata \gg 2$), the possibility of ambiguous labels increases significantly.

\noindent\textbf{Primal Objective Function} ~~ The contour lines illustrate the objective value of the primal function $\primal$, which finds the expected error for the adversarially set labels $\labels$. Since the adversarial inner optimization is a linear program, the solution jumps between vertices of the constrained polytope, making the primal expected error a piecewise linear convex function of $\learnedprobs$. 

\noindent\textbf{Adversarial Labeling} ~~ In \cref{fig:optimal}, the blue square is the minimum of the primal function, i.e., the solution to the ALL objective. This solution shows that the ideal learned model should predict $\mylabel_1$ to be positive with probability 0.18 and $\mylabel_2$ to be positive with probability 0. In the optimal state, the adversarial labeling of the examples is illustrated as the orange triangle at $(0.41, 0.23)$, i.e., the label probability vector that induces the most error for the current predicted probabilities $\learnedprobs$ that still satisfies the constraints set by $\weakvec_1$ and $\weakvec_2$. 

\noindent\textbf{Robustness to Redundant and Dependent Errors} ~~ A key feature of ALL is that it is robust to redundant and dependent errors in the weak supervision. In \cref{fig:extraweak}, we plot a variation of the setup from \cref{fig:optimal}, except we include two noisy copies of weak signal $\weakvec_2$. Since our optimal solution disagreed with weak signal $\weakvec_2$ on the most likely label for $\mylabel_1$, one might expect that adding more weak signals that agree with $\weakvec_2$ would ``outvote'' the solution and pull it to a higher probability of $\mylabel_1$ being positive. But if weak signal $\weakvec_2$ is highly correlated with weak signals $\weakvec_3$ and $\weakvec_4$, they may suffer from the same errors. Instead of these extra signals inducing a majority vote behavior on the solution, their effect on ALL is that they slightly change the feasible region of the adversarial labels, which leaves the optimum unchanged.

These two-dimensional visualizations illustrate the behavior of ALL on a simple input. In higher dimensions, i.e., when there are more examples in the training set, there is more freedom in the constraints set by each weak signal, so there will be more facets to the piecewise linear objective. 

\subsection{Optimization Approach}
\label{sec:optimization}

We use projected primal-dual updates for an augmented Lagrangian relaxation to efficiently optimize the learning objective. The advantage of this approach is that it allows inexpensive updates for all variables being optimized over, and it allows learning to occur without waiting for the solution of the inner optimization. The augmented Lagrangian form of the objective is
\begin{equation}
\begin{aligned}
\lagrangian(\params, \labels, & \lagrangevec) =
\frac{1}{\numdata} \left( \learnedprobs(\params) ^\top (1 - \labels) + (1 - \learnedprobs(\params))^\top \labels \right)\\ 
&~ - \sum_{i = 1}^\numweak \lagrange_i \left( \weakvec_i ^\top (1 - \labels) + (1 - \weakvec_i)^\top \labels - \numdata \bound_i \right)\\
&~ - \frac{\stepsize}{2} \sum_{i=1}^\numweak \left\lVert \left[ \weakvec_i ^\top (1 - \labels) + (1 - \weakvec_i)^\top \labels - \numdata \bound_i \right]_{+} \right\rVert_2^2 ~,
\end{aligned}
\end{equation}
where $[ ~ \cdot ~ ]_{+}$ is the hinge function that returns its input if positive and zero otherwise. This form uses Karush-Kuhn-Tucker (KKT) multipliers to relax the linear constraints on $\labels$ and a squared augmented penalty term on the constraint violation.

We then take projected gradient steps to update the variables $\params$, $\labels$, and $\lagrangevec$. The update step for the parameters is
\begin{equation}
\theta \leftarrow \theta - \frac{\rate_t}{\numdata} \left(\frac{\partial \learnedprobs}{\partial \params}\right)^\top \left( 1 - 2 \labels\right) ~,
\label{eq:paramupdate}
\end{equation}
where $\left(\frac{\partial \learnedprobs}{\partial \params}\right)$ is the Jacobian matrix for the classifier $\function$ over the full dataset and $\rate_t$ is a gradient step size that can decrease over time. This Jacobian can be computed for a variety of models by back-propagating through the classification computation.
The update for the adversarial labels is
\begin{equation}
\labels \leftarrow \bigg[ \labels + \rate_t \bigg( \frac{1}{\numdata}(1 - 2 \learnedprobs(\theta)) + \sum_{i=1}^\numweak \left( \lagrange_i ( 1 - 2 \weakvec_i) -  \placeholder_i ~ \right) \bigg) \bigg]_0^1 ~ ,
\label{eq:labelupdate}
\end{equation}
where
$\placeholder_i = \stepsize (1 - 2 \weakvec_i) \left[ \weakvec_i ^\top (1 - \labels) + (1 - \weakvec_i)^\top \labels - \numdata \bound_i \right]_{+}$, and $[ ~ \cdot ~ ]_0^1$ clips the label vector to the space $[0, 1]^\numdata$, projecting it into its domain. 
The update for each KKT multiplier is
\begin{equation}
\gamma_i \leftarrow \left[ \gamma_i -  \stepsize\left( \weakvec_i ^\top (1 - \labels) + (1 - \weakvec_i)^\top \labels - \numdata \bound_i \right) \right]_{+} ~ ,
\label{eq:lagrangeupdate}
\end{equation}
which is clipped to be non-negative and uses a fixed step size $\stepsize$ as dictated by the augmented Lagrangian method \cite{hestenes1969multiplier}.
These primal-dual updates for the optimization converge in our experiments. Though $\lagrangian$ is not convex with respect to $\params$, it does satisfy some of the necessary conditions for convergence derived by \citeauthor{du2018linear} (\citeyear{du2018linear}): The objective $\lagrangian$ is strongly convex in $\learnedprobs$ and $\gamma$ and concave in $\labels$, while the penalty term for the augmented Lagrangian is strongly convex. These properties may explain its convergence in practice. The full algorithm is summarized in \cref{alg:ALL}.

\begin{algorithm}[ht]
\begin{algorithmic}[1]
\REQUIRE Dataset $\Data = [\data_1, \ldots, \data_\numdata]$, learning rate schedule $\rates$, weak signals and bounds $[(\weakvec_1, \bound_1), \ldots, (\weakvec_\numweak, \bound_\numweak)]$, augmented Lagrangian parameter $\stepsize$.
\STATE Initialize $\params$ (e.g., random, zeros, etc.)
\STATE Initialize $\labels \in [0, 1]^\numdata$ (e.g., average of $\weakvec_1, \ldots, \weakvec_\numweak$)
\STATE Initialize $\lagrangevec \in \reals_{\ge 0}^\numweak$ (e.g., zeros)
\WHILE{not converged}
\STATE Update $\params$ with \Cref{eq:paramupdate}
\STATE Update $\learnedprobs$ with model and $\params$
\STATE Update $\labels$ with \Cref{eq:labelupdate}
\STATE Update $\lagrangevec$ with \Cref{eq:lagrangeupdate}
\ENDWHILE
\RETURN model parameters $\params$
\end{algorithmic}
\caption{Adversarial Label Learning}
\label{alg:ALL}
\end{algorithm}

\section{Experiments}
\label{sec:experiments}

We test adversarial label learning on a variety of datasets, comparing it with other approaches for weak supervision. In this section, we describe how we simulate domain expertise to generate weak supervision signals. We then describe the datasets we evaluated with and the compared weak supervision approaches, and we analyze the results of the experiments. 

\subsection{Simulating Weak Supervision}

\begin{figure}[tb]
\centering
\includegraphics[width=5cm]{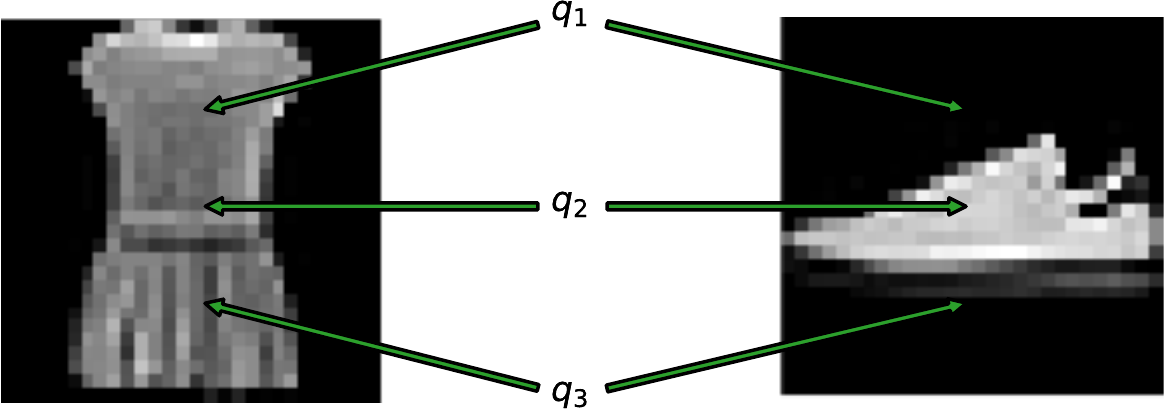}
\caption{Features used to generate weak supervision signals on Fashion-MNIST data.}
\label{fig:mnistdiagram}
\end{figure}

In practice, domain experts provide weak supervision in the form of noisy indicators or simple labeling functions. This weak supervision generates probabilities that the examples in a sample of the data belong to the positive class. Since we do not have explicit domain knowledge for the datasets used in our experiments, we generate the weak signals by training simple, one-dimensional classifiers on subsets of the data. The subset of the data used to train the weak supervision models is referred to as weak supervision data.
We train each one-dimensional weak supervision model by selecting a feature and training a one-dimensional logistic regression model using only that feature. We select the weak supervision features based on our non-expert understanding of which features could reasonably serve as indicators of the target class. For datasets whose feature descriptions are not provided, we train the weak supervision models using the first feature, middle feature, and last feature. For the Fashion-MNIST, dataset we used the pixel value at the one-quarter, center, and three-quarter locations along the vertical center line (see \cref{fig:mnistdiagram}) to build the respective weak supervision models.

We evaluate one-dimensional classifiers on the training subset, generating the weak signals $\{\weakvec_1, \ldots, \weakvec_\numweak\}$. In our first set of experiments, we measure the true error rate of each weak signal on the training subset and use that as the error bounds $\{\bound_1, \ldots, \bound_\numweak\}$. In later experiments, we set all bounds to 0.3 as an arbitrary guess.
We train weak signals from one-dimensional inputs to create realistically noisy weak signals. Training on more features could increase the predictive accuracy of the weak signals and by extension ALL, but such high-fidelity weak signals may be rare in practice. Alternatively, we chose not to hand-design weak supervision signals and bounds, because doing so could inject our own bias into this evaluation. Simulating domain expertise with a small training set provides a neutral evaluation.

\subsection{Baselines}

We compare ALL against two baseline models: a modified generalized expectation (GE) method and averaging of weak signals (AVG).

\noindent\textbf{Modified GE} ~~
GE assigns a score to the value of a model expectation. Given a conditional model distribution and a reference distribution, GE uses a score function to measure the distance between the model expectation and reference expectation. We define a modified GE method to use the label distribution conditioned on each weak signal, i.e.,
\begin{equation}
\textstyle{
\hat{p}_{\params}\left(\truelabels \vert \weakvec_k \ge 0.5 \right) = \E_{\labels}\left[\frac{1}{C_k} I(\labels) I\left(\weakvec_k \ge 0.5 \right) \right]~,}
\end{equation}
and the reference expectation is
\begin{equation}
\textstyle{
\tilde{p}\left(\truelabels \vert \weakvec_k \ge 0.5 \right) = \E_{\truelabels} \left[ \frac{1}{C_k} I(\truelabels) I\left(\weakvec_k \ge 0.5 \right) \right]~,}
\label{eq:reference}
\end{equation}
where
$\labels$ is the predicted labels and $C_k = \sum_{\weakvec_k} I\left(\weakvec_k \ge 0.5 \right) $ is a normalizing constant. We compute these reference distributions on the training subset of the data.
Our modified GE objective is then
\begin{equation}
\begin{aligned}
 \sum_{k=1}^\numweak KL\left[\tilde{p}\left(\truelabels \vert \weakvec_k \ge 0.5 \right)\Vert  \hat{p_{\params}}\left(\truelabels \vert \weakvec_k \ge 0.5 \right) \right] + \\ KL\left[\tilde{p}\left(\truelabels \vert \weakvec_k < 0.5 \right)\Vert \hat{p_{\params}}\left(\truelabels \vert \weakvec_k < 0.5 \right) \right] ~.
\end{aligned}
\end{equation}
We regularize this objective with an L2 penalty. This modified GE method is able to exploit the same information ALL is provided: the weak signals $\weakvec_1, \ldots, \weakvec_\numweak$ and the reference distributions in Eq.~\ref{eq:reference} are analogous to (though richer than) the error bounds provided to ALL.

\noindent\textbf{Averaging Baseline} ~~ 
The input to our weakly supervised learning task includes the weak supervision signals $\weakvec$, bounds $\bounds$, and the training set \emph{without labels}. A straightforward approach that a reasonable data scientist could take to this training task is to compute pseudo-labels using the weak signals. Then one can train many classifiers using a standard supervised learning approach. For the averaging method, we generate baseline models by treating the rounded average of weak signals as a label. The averaging baseline tries to mimic the aggregated weak supervision. The averaging model trains a logistic regression classifier using the average of the weak signals' predictions as labels.

\subsection{Experimental Setup}

We run experiments on nine different datasets to measure the predictive power of adversarial label learning (ALL). 
For each dataset, we generate weak supervision signals and estimate their error rates. We then compare the accuracy of the model trained by ALL against (1) the modified GE baseline, (2) the different weak supervision signals and, (3) baseline models trained by treating the average of the weak supervision signals as labels. 
We randomly split each dataset such that 30\% is used as weak supervision data, 40\% is used as training data, and 30\% is used as test data. For our experiments, we use 10 such random splits and report the mean of the results.   

In each of our experiments, we consider three different weak signals. We run ALL on the first weak signal (ALL-1), the first and second weak signals (ALL-2), or all three weak signals (ALL-3). We use the sigmoid function as our parameterized function $\function_\params$ for estimating class probabilities of ALL and GE, i.e., $\left[\function_\params(\data_j)\right]_{j=1}^\numdata =  1/(1 + \exp(-\params^T\data))= \learnedprobs_\params$.

We compare against the accuracy of GE trained using the first weak signal (GE-1), the first and second weak signals (GE-2), or all three weak signals (GE-3). We also compare directly using the individual weak signals as the classifier (WS-1, WS-2, and WS-3). And finally, we train models to mimic the average of the first weak signal (AVG-1), the first and second weak signals (AVG-2), and all three weak signals (AVG-3). \Cref{tab:test} shows the mean accuracies obtained by running ALL on the different datasets.

\begin{table*}[ht]
\centering       
\resizebox{\textwidth}{!}{
\begin{tabular}{l rrr rrr rrr rrr}
\toprule
Dataset & ALL-1 &ALL-2 &ALL-3 &GE-1 & GE-2 &GE-3 & AVG-1 &AVG-2 &AVG-3 &WS-1 &WS-2 &WS-3  \\
\midrule
Fashion MNIST (DvK) &\textbf{0.998} &\textbf{0.995} &\textbf{0.996} &0.975 &0.972 &0.977 &0.506 &0.743 &0.834 &0.508 &0.750 &0.644 \\ 
 
Fashion MNIST (SvA) &\textbf{0.923} &\textbf{0.922} &\textbf{0.924} &0.501 &0.500 &0.500 &0.561 &0.568 &0.719 &0.562 &0.535 &0.688 \\

Fashion MNIST (CvB) &0.795 &\textbf{0.831} &\textbf{0.840} &0.497 &0.499 &0.500 &0.577 &0.697 &0.740 &0.587 &0.684 &0.643 \\ 
 
Breast Cancer &\textbf{0.942} &\textbf{0.944} &\textbf{0.945} &\textbf{0.936} &\textbf{0.936} &\textbf{0.935} &0.889 &0.885 &0.896 &0.871 &0.804 &0.915  \\

OBS Network &\textbf{0.717} &\textbf{0.718} &\textbf{0.719} &0.708 &0.701 &0.698 &\textbf{0.724} &\textbf{0.723} &0.698 &\textbf{0.721} &0.715 &0.692  \\

Cardiotocography &0.803 &0.803 &0.803 &0.824 &0.675 &0.633 &\textbf{0.942} &\textbf{0.947} &\textbf{0.942} &\textbf{0.946} &0.602 &0.604  \\

Clave Direction &0.646 &\textbf{0.837} &0.746 &0.646 &0.796 &0.772 &0.646 &0.645 &0.707 &0.646 &0.648 &0.625 \\

Credit Card &\textbf{0.697} &\textbf{0.696} &\textbf{0.697} &\textbf{0.695} &0.460 &0.424 &0.660 &0.662 &0.607 &0.659 &0.572 &0.557 \\

Statlog Satellite &0.470 &0.933 &0.936 &0.521 &\textbf{0.987} &\textbf{0.992} &0.669 &0.926 &0.916 &0.660 &0.775 &0.880  \\

Phishing Websites &\textbf{0.896} &\textbf{0.895} &\textbf{0.895} &\textbf{0.898} &\textbf{0.894} &0.870 &0.846 &0.807 &0.846 &0.846 &0.700 &0.585  \\

Wine Quality &0.572 &\textbf{0.662} &0.623 &0.455 &0.427 &0.454 &0.570 &0.573 &0.555 &0.571 &0.596 &0.570  \\

\bottomrule
\end{tabular}}
\caption{Test accuracy of ALL and baseline models on different datasets. The best performing methods that are not statistically distinguishable using a two-tailed paired t-test (p = 0.05) are boldfaced.}
\label{tab:test}
\end{table*}

\subsection{Datasets}

We describe the datasets used in the experiments below.

\noindent\textbf{Fashion-MNIST} ~~
The Fashion-MNIST dataset \cite{xiao2017fashion} represents an image-classification task where each example is a $28 \times 28$ grayscale image. The images are categorized into 10 classes of clothing types. Each class contains 6,000 training examples and 1,000 test examples. We consider the binary classification between three pairs of classes: dresses/sneakers (DvK), sandals/ankle boots (SvA), and coats/bags (CvB).

\noindent\textbf{Breast Cancer} ~~
The task in this dataset is to diagnose if the breast cell nuclei are from a malignant (positive) or benign (negative) case of breast cancer \cite{uci,street1993nuclear}. We use the mean radius of the nucleus (WS-1), the radius standard error (WS-2), and worst radius (WS-3) of the cell nucleus as features to train the three different weak supervision models. The dataset contains 569 samples.

\noindent\textbf{OBS Network} ~~
The classification task for the Burst Header Packet Flooding Attack Detection dataset is to detect network nodes based on their behavior, identifying whether they should be blocked for potentially malicious behavior \cite{rajab2016countering}. 
We use the percentage of flood per node (WS-1), average packet drop rate (WS-2), and utilized bandwidth (WS-3) as features to train the weak signals. The original dataset contains four classes, so we select the two classes with the most examples, resulting in a total of 795 examples.

\noindent\textbf{Cardiotocography} ~~
The task for this dataset is to classify fetal heart rate using uterine contraction features on cardiotocograms classified by expert obstetricians \cite{ayres2000sisporto}.
The original dataset contains 10 classes, we select the most common two classes, resulting in a total of 963 examples.
We use accelerations per second (WS-1),  mean value of long-term variability (WS-2), and histogram median (WS-3) as features to train the weak signals.

\noindent\textbf{Clave Direction} ~~
The task for the Firm Teacher Clave Direction dataset is to classify the clave direction from rhythmic patterns \cite{vurkacc2011clave}.
The original dataset contains four classes, so we select the two most common classes, resulting in a total of 8,606 examples.
We use the first (WS-1),  middle (WS-2), and last (WS-3) features to train the weak signals.

\noindent\textbf{Credit Card} ~~
The Statlog German Credit Card dataset task is to classify people described by a set of attributes as good or bad credit risks \cite{uci}.
We use the status of an existing checking account (WS-1),  installment rate in percentage of disposable income (WS-2), and amount of existing credit at the bank (WS-3) as features to train the weak signals. The dataset contains 1,000 samples.

\noindent\textbf{Statlog Satellite} ~~
The task of the Statlog dataset is to predict soil class given the multi-spectral values of pixels in 3x3 neighborhoods of satellite images \cite{uci}.
The original dataset contains seven classes of soil samples, so we select the two most common classes, resulting in a total of 3,041 examples.
We use the first (WS-1),  middle (WS-2), and last (WS-3) features to train the weak signals.

\noindent\textbf{Phishing Websites} ~~
The task is to identify phishing websites using different web attributes \cite{mohammad2012assessment}.
The dataset contains 11,055 samples.
We use the URL of the anchor (WS-1),  web traffic (WS-2), and Google index (WS-3) as features to train the weak signals.

\noindent\textbf{Wine Quality} ~~
The task is to classify the quality of wine using physiochemical attributes of the wine \cite{cortez2009modeling}.
The original dataset contains seven classes, so we select the two classes with the most examples, resulting in a total of 4974 examples.
We use fixed acidity (WS-1),  density (WS-2), and pH (WS-3) as features to train the weak signals.

\begin{figure}[tb]
\centering
\includegraphics[width=7cm]{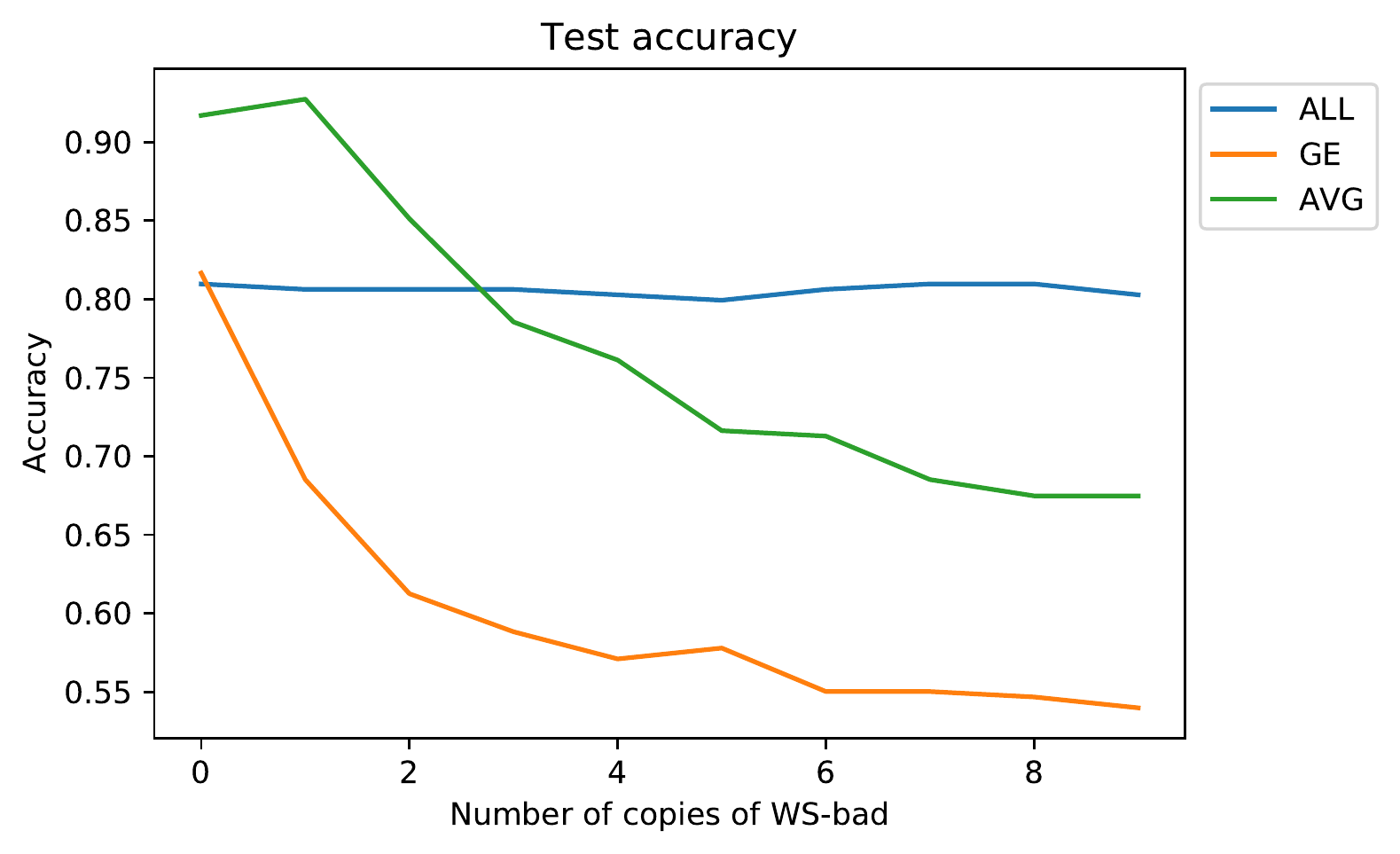}
\caption{Performance of the methods using one good weak signal and repeated erroneous weak signals.}
\label{fig:dependent}
\end{figure}

\subsection{Learning with True Bounds}

Our first experiments allow ALL to use the error bounds computed on the training set. 
\Cref{tab:test} shows the accuracies of the models evaluated on the held-out test sets of each task. ALL trains models that perform significantly better than the weak signals and the baselines on the test data. The AVG baselines perform better with an increasing number of weak signals, but their best accuracy score on most datasets is significantly worse than that of ALL. ALL trains a robust model and is able to learn using noisy weak signals. Despite the fact that the weak signals on the Fashion MNIST dataset have rather low accuracy, ALL trained with these signals is able to achieve high accuracy. The GE method only significantly outperforms ALL on the Statlog Satellite dataset, and nevertheless ALL still achieves a high accuracy score. The main failure case is the cardiotocography task, in which the AVG baseline outperforms both GE and ALL. However, in this task and others, we observe that ALL performs well even when the weak signals make dependent errors, while the baseline methods suffer as more signals with dependent errors are introduced. We study this concept further in the next experiment.

\begin{table*}[ht]
\centering       
\resizebox{\textwidth}{!}{
\begin{tabular}{l rrr rrr rrr rrr}
\toprule
Dataset & ALL-1 &ALL-2 &ALL-3 &GE-1 & GE-2 &GE-3 & AVG-1 &AVG-2 &AVG-3 &WS-1 &WS-2 &WS-3  \\
\midrule
Fashion MNIST (DvK) &\textbf{0.998} &\textbf{0.995} & \textbf{0.996} &0.975 &0.972 &0.977 &0.506 &0.743 &0.834 &0.508 &0.750 &0.644 \\ 
 
Fashion MNIST (SvA) &\textbf{0.895} &0.825 &\textbf{0.901} &0.501 &0.500 &0.500 &0.561 &0.568 &0.719 &0.562 &0.535 &0.688  \\

Fashion MNIST (CvB) &\textbf{0.810} &\textbf{0.805} &\textbf{0.802} &0.497 &0.499 &0.500 &0.577 &0.697 &0.740 &0.587 &0.684 &0.643  \\ 
 
Breast Cancer &\textbf{0.940} &\textbf{0.941} &\textbf{0.944} &\textbf{0.936} &\textbf{0.936} &\textbf{0.935} &0.889 &0.885 &0.896 &0.871 &0.804 &0.915  \\

OBS Network &\textbf{0.719} &\textbf{0.719} &\textbf{0.722} &0.708 &0.701 &0.698 &\textbf{0.724} &\textbf{0.723} &0.698 &\textbf{0.721} &0.715 &0.692  \\

Cardiotocography &0.805 &0.794 &0.657 &0.824 &0.675 &0.633  &\textbf{0.942} &\textbf{0.947} &\textbf{0.942} &\textbf{0.946} &0.602 &0.604 \\

Clave Direction &0.646 &\textbf{0.854} &0.727 &0.646 &0.796 &0.772 &0.646 &0.645 &0.707 &0.646 &0.648 &0.625 \\

Credit Card &\textbf{0.696} &0.671 &0.610 &\textbf{0.695} &0.460 &0.424 &0.660 &0.662 &0.607 &0.659 &0.572 &0.557 \\

Statlog Satellite &0.493 &\textbf{0.983} &\textbf{0.982} &0.521 &\textbf{0.987} &\textbf{0.992} &0.669 &0.926 &0.916 &0.660 &0.775 &0.880  \\

Phishing Websites &\textbf{0.899} &0.835 &0.853 &\textbf{0.898} &\textbf{0.894} &0.870 &0.846 &0.807 &0.846 &0.846 &0.700 &0.585  \\

Wine Quality &0.566 &0.603 &\textbf{0.694} &0.455 &0.427 &0.454 &0.570 &0.573 &0.555 &0.571 &0.596 &0.570  \\

\bottomrule
\end{tabular}}
\caption{Test accuracy of ALL and baseline models on different datasets using fixed bounds. The best performing methods that are not statistically distinguishable using a two-tailed paired t-test (p = 0.05) are boldfaced.
We replicate the baseline results from the previous experiments for convenience; they are unaffected by the change in error bound.}
\label{tab:fixed_test} 
\end{table*}

\begin{figure*}[tbh]
    \centering
    \begin{subfigure}[b]{0.40\textwidth}
        \includegraphics[width=\textwidth]{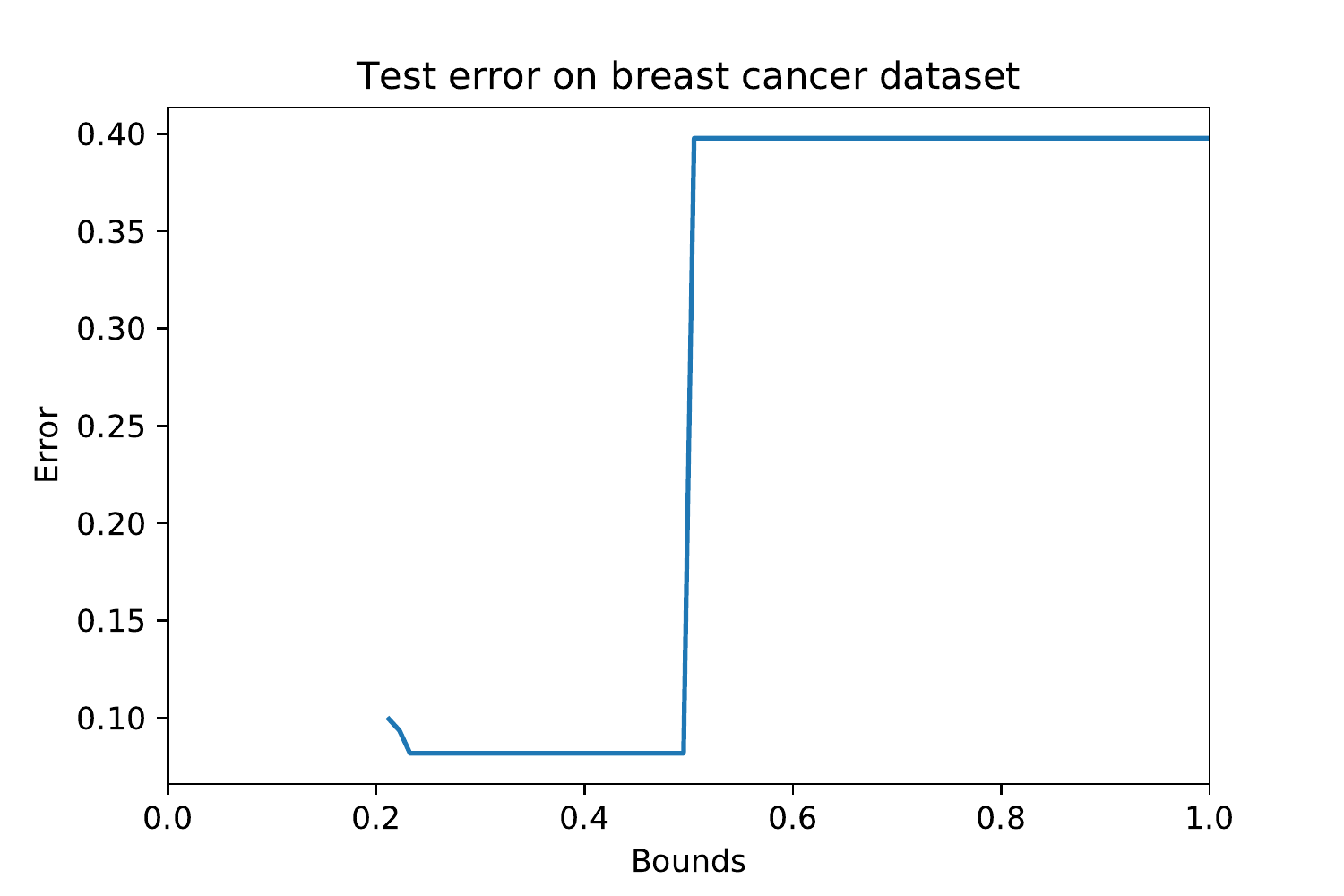}
        \label{fig:bc_error}
    \end{subfigure}
    ~ 
    \begin{subfigure}[b]{0.40\textwidth}
        \includegraphics[width=\textwidth]{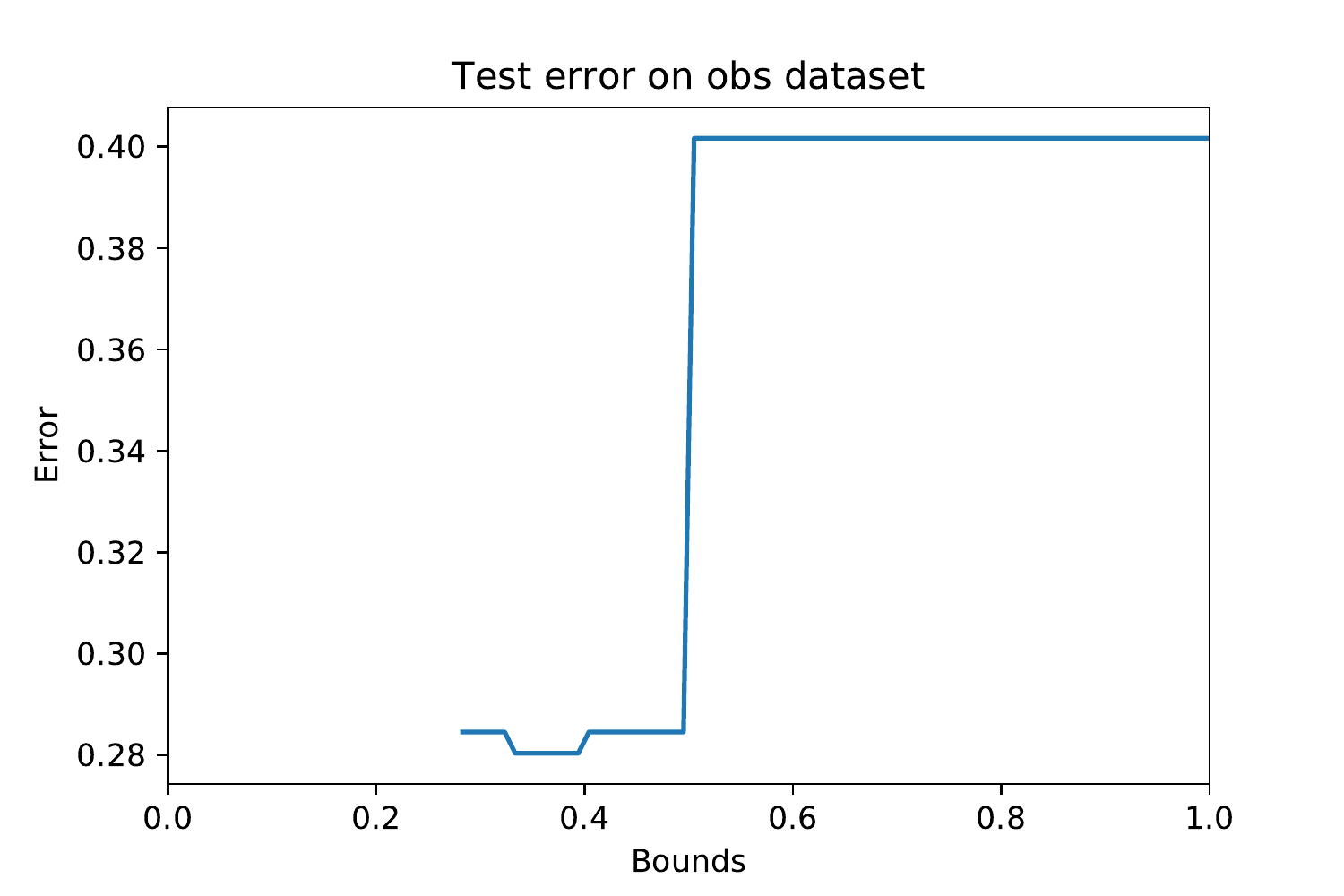}
        \label{fig:obs_error}
    \end{subfigure}
    \caption{Illustrations showing the error of the model (ALL-3) when run with different fixed bounds between 0 and 1. Small bound values make infeasible constraints that prevent convergence, and are not plotted here. }\label{fig:error}
\end{figure*}

\subsection{Robustness against Dependent Errors}
\label{dependent_error}
We observed from our test results that unlike the baselines, ALL learns a robust model that performs well even in the presence of low-quality weak signals. We isolate this concept using two weak signals from the cardiotocography task, a high-quality weak signal (WS-good) and a low-quality weak signal (WS-bad). We consider the scenario where the low-quality signal (WS-bad) is copied multiple times in the weak supervision. We train the models with WS-good and a varying number of copies of WS-bad. We evaluate the performance of the models on each experiment using the test data. \Cref{fig:dependent} plots the accuracy of the models under these settings. In the presence of multiple dependent erroneous weak signals, ALL's performance is relatively stable while the baseline accuracies get worse as the poor performing weak signal is repeated. The accuracy of AVG steadily degrades, while GE declines steeply to random performance. 

\subsection{Learning with Fixed, Incorrect Bounds}

Instead of using the true training error as the bounds, we consider a more realistic scenario in which the experts are less precise about their error estimates. In practice, the true error rate may be difficult to estimate, so these experiments will validate whether our approach continues to work well when these bounds are inaccurate. We use a fixed upper bound of $\bound_1 = \bound_2 = \bound_3 = 0.3$ and report the performance of the ALL model and baselines in this setting.

\Cref{tab:fixed_test} shows the accuracies obtained by the methods using the fixed bounds.
The accuracy scores from the Statlog Satellite datasets are marginally higher than the results from the previous experiments, which used the true error rate (see \cref{tab:test}), making it's performance statistically indistinguishable compared to GE. 

While we arbitrarily chose a fixed bound of 0.3, we also tried various values of the bound, finding that ALL is not too sensitive to variations of this parameter. The only real challenge in setting this parameter is that when the bound is small enough, the problem becomes infeasible. See \cref{fig:error}.

\section{Conclusion}
\label{sec:conclusion}

We introduced adversarial label learning (ALL), a method to train robust classifiers when access to labeled training data is limited. ALL trains a model without labeled data by making use of weak supervision to minimize the error rate for adversarial labels, which are subject to constraints defined by the weak supervision. We demonstrated that our method is robust against weak supervision signals that make dependent errors. Our experiments confirm that ALL is able to learn models that outperform the weak supervision and baseline models. ALL is also capable of directly training classifiers to mimic the weak supervision.

While our contribution is a significant methodological advance, there are several directions we hope to explore in our future work.
We focused on training binary classifiers, but the principles underlying our method should extend to multi-class, regression, and even structured-output settings. Our algorithm requires reasoning over the entire training dataset, so we will explore ideas for scalability such as stochastic variations of our optimization procedure. 

\bibliographystyle{aaai}
\bibliography{arachie.bib}

\end{document}